%% file: antqa.tex
\pdfoutput=1
\documentclass[sigconf]{acmart}  
\copyrightyear{2022} 
\acmYear{2022} 
\setcopyright{acmcopyright}\acmConference[WSDM '22]{Proceedings of the Fifteenth ACM International Conference on Web Search and Data Mining}{February 21--25, 2022}{Tempe, AZ, USA}
\acmBooktitle{Proceedings of the Fifteenth ACM International Conference on Web Search and Data Mining (WSDM '22), February 21--25, 2022, Tempe, AZ, USA}
\acmPrice{15.00}
\acmDOI{10.1145/3488560.3498378}
\acmISBN{978-1-4503-9132-0/22/02}

\usepackage{booktabs} 
\usepackage{graphicx}  
\usepackage{CJK}
\usepackage{multirow}
\usepackage{subfigure}
\usepackage{makecell}
\usepackage{xspace}
\usepackage{colortbl}
\usepackage{xcolor}

\newcommand{\eg}{\emph{e.g.,}\xspace}

\newcommand{\ignore}[1]{}

\newcommand{\dubbelop}{$^{\blacktriangle}$}

\newcommand{\dubbelneer}{$^{\blacktriangledown}$}

\AtBeginDocument{%
  \providecommand\BibTeX{{%
    \normalfont B\kern-0.5em{\scshape i\kern-0.25em b}\kern-0.8em\TeX}}}

\begin{document}
\fancyhead{}
\title{HeteroQA: Learning towards Question-and-Answering \\ through Multiple Information Sources \\ via Heterogeneous Graph Modeling} 


\author{Shen Gao\textsuperscript{1}, Yuchi Zhang\textsuperscript{2}, Yongliang Wang\textsuperscript{2}, Yang Dong\textsuperscript{2}, Xiuying Chen\textsuperscript{5}, Dongyan Zhao\textsuperscript{1}and Rui Yan\textsuperscript{3,4}}
\authornote{Corresponding Author: Rui Yan (ruiyan@ruc.edu.cn) and Dongyan Zhao (zhaody@pku.edu.cn)}
\affiliation{
  \institution{
    \textsuperscript{1} Wangxuan Institute of Computer Technology, Peking University \\
    \textsuperscript{2} Ant Group \\
    \textsuperscript{3} Gaoling School of Artificial Intelligence, Renmin University of China \\
    \textsuperscript{4} Beijing Academy of Artificial Intelligence \\
  \textsuperscript{5} King Abdullah University of Science and Technology \\
  }
  \country{}
}
\email{{shengao, zhaody}@pku.edu.cn, {yuchi.zyc, yongliang.wyl, doris.dy}@alibaba-inc.com}
\email{xiuying.chen@kaust.edu.sa, ruiyan@ruc.edu.cn}

\renewcommand{\shortauthors}{Shen and Yuchi, et al.}

\begin{abstract}
  Community Question Answering (CQA) is a well-defined task that can be used in many scenarios, such as E-Commerce and online user community for special interests. 
  In these communities, users can post articles, give comment, raise a question and answer it.
  These data form the heterogeneous information sources where each information source have their own special structure and context (comments attached to an article or related question with answers).
  Most of the CQA methods only incorporate articles or Wikipedia to extract knowledge and answer the user's question.
  However, various types of information sources in the community are not fully explored by these CQA methods and these multiple information sources (MIS) can provide more related knowledge to user's questions.
  Thus, we propose a question-aware heterogeneous graph transformer to incorporate the MIS in the user community to automatically generate the answer.
  To evaluate our proposed method, we conduct the experiments on two datasets: $\text{MSM}^{\text{plus}}$ the modified version of benchmark dataset MS-MARCO and the AntQA dataset which is the first large-scale CQA dataset with four types of MIS\footnote{\url{https://github.com/gsh199449/HeteroQA}}.
  Extensive experiments on two datasets show that our model outperforms all the baselines in terms of all the metrics.
\end{abstract}



\maketitle

\input{intro}
\input{related}
\input{dataset}

\input{model}

\input{exp}
\input{conclusion}

\section*{Acknowledgments}

We would like to thank the anonymous reviewers for their constructive comments. 
We would also like to thank Nan Su for her help on this project.
This work was supported by the National Key Research and Development Program of China (No. 2020AAA0106600), the National Science Foundation of China (NSFC No. 62122089 and No. 61876196).
Rui Yan is supported as a young fellow at Beijing Academy of Artificial Intelligence (BAAI).

\bibliographystyle{ACM-Reference-Format}
\bibliography{references}

\end{document}

%% file: intro.tex
\section{Introduction}

\begin{figure*}
    \centering
    \includegraphics[width=0.8\textwidth]{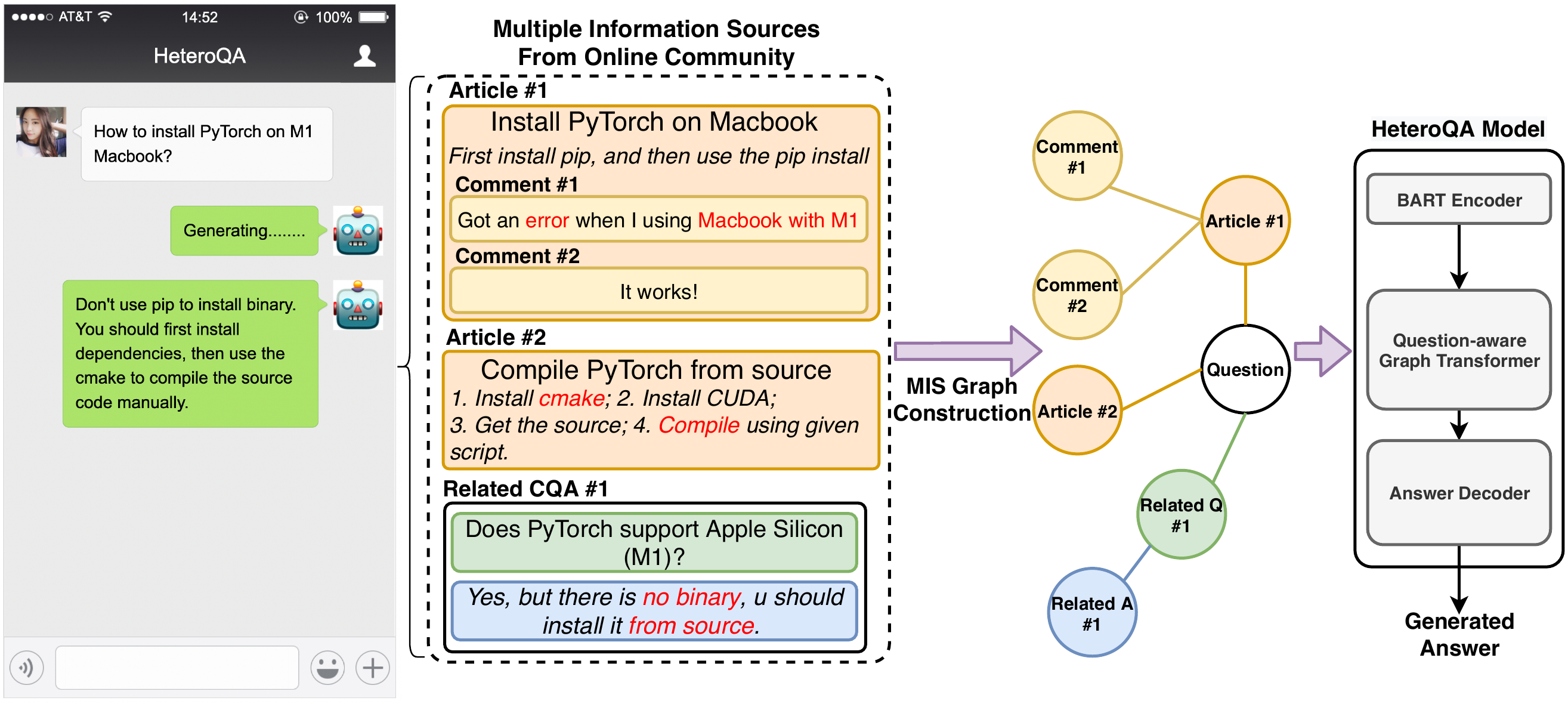}
    \caption{Example of incorporating heterogeneous information sources in community question answering.}
    \label{fig:intro-case}
\end{figure*}

Online user community is widely adopted in many scenarios, such as software user communities (pytorch forum), interest groups (sub-reddit in Reddit and XueQiu), general communities (Quora and ZhiHu) and E-commerce websites (Taobao and AntFortune). 
To respond to user's queries timely and improve the experience, researchers propose the Community Question Answering (CQA)~\cite{Ruckle2019Improved,Fu2020Recurrent,Wu2018Question} which first reads the community article or Wikipedia and answers the user's question.
Most previous works focus on multi-hop retrieving~\cite{Min2020Knowledge} or modeling the fine-grained interaction between question and documents~\cite{Wang2019Document}.

However, most of the previous CQA models use a single information source (\eg articles or knowledge base) and some of the information comes from an external source (\eg Wikipedia), and it cannot cover the various aspects of user questions.
In fact, there is various information in the online user community which has not been fully explored by existing CQA methods, such as article comments and related question-answer pairs.
Moreover, the data structure and context of these multiple information sources (MIS) are very different: comments are associated with plain text articles and the related question-answering are pair-wise text data. 
Thus, these multiple information sources form a heterogeneous knowledge source which can help the CQA model improve the answer quality, as shown in Figure~\ref{fig:intro-case}.
The reason lies in that these information sources are supplements to the community articles and these sources from the same community are more relevant to the question than the retrieved external Wikipedia.
Therefore, in this paper, we propose to incorporate four different types of MIS into the community question answering: (1) user articles; (2) article comments; (3) related questions; (4) their answers.

Since MIS contains heterogeneous information form, we cannot simply treat them as plain text or model each type of MIS equally and independently, and each MIS should be modeled with its own structure and context.
For example, the article comment cannot be understood without the semantic context of the article, and we should build an explicit connection between the comment and the article.
To incorporate MIS into answer generation, there are two challenges that should be tackled: 
(1) How to unify the heterogeneous multiple information sources and extract useful knowledge according to the user question; 
(2) How to incorporate the knowledge learned from the heterogeneous MIS into the final answer generation process.
Thus, we propose the \textbf{Hetero}geneous graph based \textbf{Q}uestion \textbf{A}nswering model (HeteroQA).
We first construct a heterogeneous graph using the different types of sources as a different type of node, and the graph construction method is shown in Figure~\ref{fig:intro-case}.
Then, we propose a question-aware graph reasoning method on the MIS heterogeneous graph to extract question-related knowledge from MIS.
Finally, we generate the answer by dynamically incorporating graph node based on a BART-based~\cite{Lewis2020BART} pretrained text generation framework.
To demonstrate the effectiveness of the proposed HeteroQA model, we conduct experiments on two datasets: (1) $\text{MSM}^{\text{plus}}$ dataset which is a modified version of the public benchmark CQA dataset MS-MARCO~\cite{Campos2016MSMA} and (2) a large-scale CQA dataset AntQA with four types of MIS that we released in this paper.
Experiments demonstrate that HeteroQA outperforms all the strong baselines on both datasets.

\noindent To sum up, our contributions can be summarized as follows:

$\bullet$ We propose to incorporate MIS in the online user community to facilitate the CQA model.

$\bullet$ To leverage this heterogeneous information source, we propose the Heterogeneous graph-based Question Answering model (HeteroQA).

$\bullet$ Experiments show that our HeteroQA outperforms all the strong baselines on two datasets: MS-MARCO and AntQA which is a large-scale CQA dataset with MIS proposed in this paper.

%% file: related.tex
\section{Related Work}\label{sec:related}

\textbf{Community Question Answering.}
The current CQA task aims to automate the process of finding appropriate answers to questions in a community-created discussion forum~\cite{Nakov2016SemEval2016T3}. 
Recent researches~\cite{Zhang2017Attentive,Xie2020Attentive} often regard the CQA task as a text-matching problem and have proposed various deep learning networks to learn the semantic representation of question-answer pairs.

Different from the answer selection and span extraction methods, researchers propose the generative-based CQA methods~\cite{Deng2020Joint} which can generate the natural language answers.
Answering the custom question on E-commerce website is an important CQA application which has been addressed by many researchers~\cite{Gao2019ProductAware,Deng2020Opinionaware,Gao2021Meaningful}, and most of these methods tackle this task as a knowledge-grounded text generation that leverages the custom reviews and product attributes to generate the answer.

However, existing CQA methods do not consider the MIS content in the community, which can provide more related knowledge than using the articles along to answer the question.

\textbf{Heterogeneous Graph Network.}
Heterogeneous graph is widely used in many structure data modeling tasks: recommendation~\cite{Liu2020A,Jiang2018Crosslanguage,Fan2019Metapathguided}, social network modeling~\cite{Wu2019Graph,Fan2019Graph,Yang2020Rumor}, text summarization~\cite{Wang2020Heterogeneous}, and knowledge-graph based tasks~\cite{Jiang2020Reasoning,Wu2019RelationAware}.
In the CQA task, most of the researchers employ the homogeneous graph to model the contextual information, and only a few CQA methods leverage the heterogeneous graph model.
\citet{Hu2020Multi,Hu2019Hierarchical} propose to use the graph to model the multi-modal content of questions and answers with a graph neural network.
\citet{Sun2020EndCold} propose a graph-based model to route the newly posted questions to potential answers, which models the relationship between question, tags, and users using a heterogeneous graph.
However, how to model the different types of MIS text using heterogeneous graph has not been studied in previous research, and most of the previous methods only retrieve existing answers or select answer from pre-defined candidates.
In this paper, we focus on incorporating MIS from the user community and generate a proper answer from scratch.

\textbf{Text Generation Pretraining.}
Recently, large-scale language models based on transformer~\cite{Vaswani2017Attention} pretrained with mask language modeling~\cite{Devlin2019BERT} or text infilling~\cite{Lewis2020BART} have been explored further advanced the state-of-the-art on many language understanding and generation tasks~\cite{Li2021Stylized}.
We focus on improving the factual correctness and language fluency of the CQA model by using these pretraining models.

%% file: dataset.tex
\section{AntQA Dataset}\label{sec:dataset}

In this paper, we propose a new large-scale community question-and-answering (CQA) dataset named AntQA, and the biggest difference compared to existing CQA datasets lies in that AntQA contains multiple types of information sources to support the question-answering task.
We will first introduce the data collection methods of AntQA, and then give some statistics of the QA pair and the multiple information sources.

\begin{CJK*}{UTF8}{gbsn}
    \begin{table}[t]
    \centering
    \caption{Examples of the community QA and MIS.}
    \label{tab:data-case}
    \resizebox{1\columnwidth}{!}{
    \begin{tabular}{l}
    \toprule
    \multicolumn{1}{p{1\columnwidth}}{\textbf{Question}: 在支付宝上借款怎么还款 (How to conduct repayment on Alipay?)} \\
    \multicolumn{1}{p{1\columnwidth}}{\textbf{Answer}: 可以在支付宝里面进行还款 (Use the Alipay mobile app to conduct the repayment.)} \\
    \multicolumn{1}{p{1\columnwidth}}{\textbf{MIS (Related QA)}: \textbf{Q}: 支付宝借呗怎么还？(How to repay for ``Jiebei'' (Ant Microloan)?) \textbf{A}: 进入支付宝借呗页面，就可以看到去借钱和去还钱，点击还款就可以看到还款数目，将足够的资金提前存入支付宝的余额 (Login the ``Jiebei'' page, and you can see ``Borrow money'' and ``Repayment''. Click on the ``Repayment'' button, and then you can see the amount you should pay back. Deposit enough money in your Alipay balance account.)} \\
    \multicolumn{1}{p{1\columnwidth}}{\textbf{MIS (Related Article)}: 借呗提前还款要收手续费吗？不需要，借呗提前还款暂时没有手续费，大家直接在借呗页面操作即可。登录手机支付宝，点击右下角，在借呗首页选择，按页面提示操作。(Do I have to charge a handling fee for early repayment? No, there is no handling fee for repayment in ``Jiebei''. So you can directly conduct repayment on the ``Jiebei''. First, log in to Alipay mobile version, click on the lower right corner, select on the home page of ``Jiebei'', and follow the instructions on the page.)} \\ \multicolumn{1}{p{1\columnwidth}}{\textbf{MIS (Related Article)}: 支付宝信用卡还款从3月26日起超出2000元部分将收取0.1\%手续费的消息，自昨天起在各大平台传得沸沸扬扬。相信大家其实都已经早有了心理准备。 (The news that Alipay credit card repayments will be charged a 0.1\% handling fee for more than 2,000 yuan from March 26 has been circulated on major platforms since yesterday. I believe that everyone has already been mentally prepared.) \textbf{Comments}: \#1. 谢谢分享！ (Thanks for sharing!) \#2. 自从微信收费以来，我一直是通过支付宝来还款 (Since WeChat charges the handling fee, I have always used Alipay to repay)} \\
    \bottomrule
    \end{tabular}
    }
    \end{table}
\end{CJK*}

\subsection{Data Collection}

To construct a community question-answering dataset, we collect the textual data from the AntFortune~\footnote{https://antfortune.com} which is one of the largest financial forums in China.
In this financial forums, users in community disscuss about the stock and mutual fund by publishing articles or sending tweets about their analysis of stock market.
Other users can give comments to the articles and tweets.
Most importantly, like most of online communities, users can pose a question and answer others question.
In this community, practitioners in the financial industry can obtain verification by showing their credentials to the administration of the online community.
Thus, we collect the QA pairs which are answered by the verification users.

To build the multiple information sources (MIS), we collect four types of information sources including: (1) user articles; (2) article comments; (3) related questions and (4) its answers.
For a community QA pair, we use the question as the query to retrieve the related question-answer from all the training QA samples using BM25 algorithm, and we use a popular information retrieval system ElasticSearch\footnote{\url{https://www.elastic.co/cn/elasticsearch/}} to implement distributed BM25 algorithm.
And we also retrieve the user articles and its comments using the same method as the related QA pairs.

Table~\ref{tab:data-case} shows a randomly sampled example data from the test set of AntQA.
In this example, we show the question, answer and three MIS item which contains one related QA pair and two user articles.
And the second article has two user comments.
This example question asks about the usage of Alipay app, which topic is not a popular topic disscussed by user articles.
And how to repayment is a very common question espacially for new users of Alipay, thus we can find the useful knowledge from the related QA pairs in MIS.
However, not all the questions are talked about the simple topics, for example, other questions ask the trend of stock market should be answered by incroporating the user articles.

According to the user agreement and privacy policy of this online community, we are authorized to use and release the anonymous CQA and MIS data for research.

To verify the generalization of our proposed method in this paper, we also propose to modify a benchmark dataset MS-MARCO~\cite{Campos2016MSMA} as a new dataset $\text{MSM}^{\text{plus}}$.
We employ the information retrieval system ElasticSearch to retrieve several similar QA pairs for each CQA sample and forming the new CQA dataset $\text{MSM}^{\text{plus}}$ with related QA pairs as MIS.
Since the MS-MARCO leaderboard is closed, we use the development set as the test set for $\text{MSM}^{\text{plus}}$.
We will also introduce the statistics of $\text{MSM}^{\text{plus}}$ in the next section for comparison with AntQA.

\subsection{Key Statistics}

\begin{table}[!t]
    \caption{Dataset Statistics. The MIS are associated with each QA sample. The number of articles denotes the retrieved articles for a QA sample. The number of comments denotes the total number for the associated articles of a QA sample.}
    \label{tbl:data-statistics}
    \centering
    \begin{tabular}{lll}
     \toprule
     & AntQA & $\text{MSM}^{\text{plus}}$ \\
     \midrule 
     \# of training samples & 376948 & 387774 \\
     \# of test samples & 2409 & 49230 \\
     \# of validation samples & 1920 & - \\
     Avg. words of question & 25.88 & 5.77 \\
     Avg. words of answer & 83.89 & 16.27 \\
     Avg. number of articles & 29.99 & 9.98 \\
     Avg. sentences of articles & 3.87 & 5.04 \\
     Avg. words of articles & 297.01 & 60.49 \\
     Avg. number of comments & 11.63 & - \\
     Avg. words of comments & 18.83 & - \\
     Avg. number of related QA & 9.01 & 6.37 \\
     \bottomrule
    \end{tabular}
\end{table}

We list some basic statistics for AntQA and $\text{MSM}^{\text{plus}}$ in Table~\ref{tbl:data-statistics}.
AntQA and $\text{MSM}^{\text{plus}}$ contain comparable number of training samples, and AntQA contains 376948 training samples, 1920 validation samples, and 2409 test samples.
The AntQA contains more than three times supported articles than $\text{MSM}^{\text{plus}}$ dataset, and AntQA has more related QA pairs than $\text{MSM}^{\text{plus}}$.
In this paper, we use the character as the basic unit of Chinese and use the word unit for English.
And the lengths of QA and articles in $\text{MSM}^{\text{plus}}$ dataset are also shorter than the AntQA dataset.

To verify the correctness of the retrieved MIS in AntQA, we also conduct the human evaluation using scores 1-3 (3 means the MIS has completely correct information to answer the question).
The average score is 2.12 and there are 72\% samples that obtain a score equal to or greater than 2 which demonstrates the majority of MIS can support the model to give a correct answer.

%% file: model.tex
\section{Problem Formulation}\label{sec:problem-formulation}

Given a user question $q = \{w^q_1, \cdots, w^q_{L_q}\}$ with $L_q$ tokens and $T$ types of related MIS $u = \{u^1, \cdots, u^T\}$ which is retrieved from the database using the question as the query, our goal is to generate an answer $a = \{w^a_1, \cdots, w^a_{L_a}\}$ with $L_a$ tokens according to the knowledge learned from the related MIS.
In each type of MIS $u^i$, it contains $L^i_u$ documents $u^i = \{d^i_1, \cdots, d^i_{L^i_u}\}$, and the document $d^i_j = \{d^i_{j,1}, \cdots, d^i_{j,L_d}\}$ contains $L_d$ tokens. 
In this paper, we retrieve four types of related MIS: (1) user articles; (2) article comments; (3) related questions and (4) its answers.
Finally, we use the difference between generated answer $\hat{a}$ and the ground truth answer $a$ as the training objective.

\begin{figure*}
    \centering
    \includegraphics[scale=0.55]{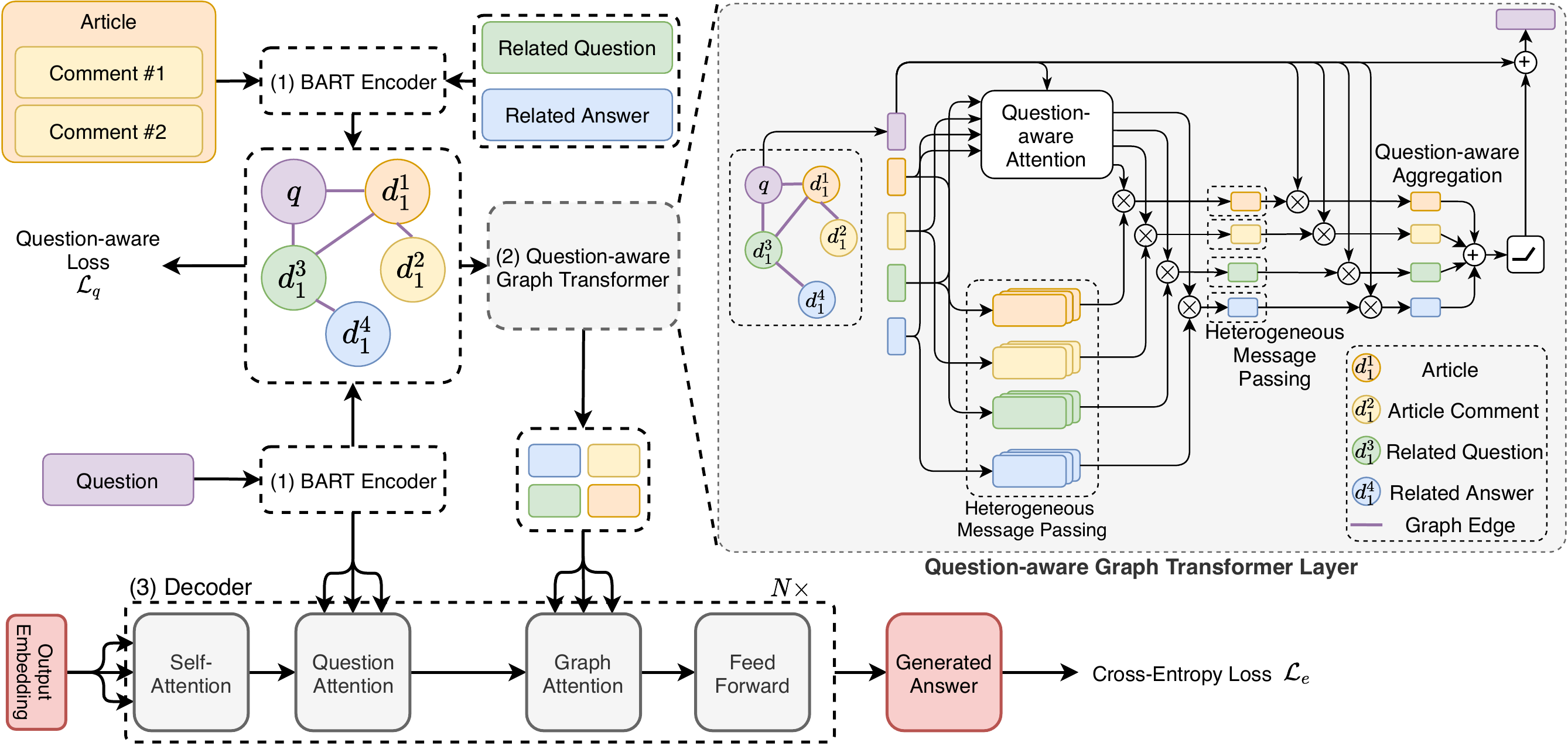}
    \caption{
        Overview of HeteroQA. We divide our model into three parts: (1) BART-based Node Encoder; (2) Question-aware Graph Transformer; (3) Answer Decoder.
    }
    \label{fig:model}
\end{figure*}

\section{HeteroQA Model}\label{sec:model}

\subsection{Overview}
In this section, we introduce the \textbf{Hetero}geneous graph based \textbf{Q}uestion \textbf{A}nswering model (HeteroQA).
An overview of HeteroQA is shown in Figure~\ref{fig:model}, which has three main parts:

\noindent $\bullet$ \textbf{Node Encoder} employs the BART~\cite{Lewis2020BART} to encode the question and each node text, and then uses the output hidden state of BART encoder as the initial graph node representation.

\noindent $\bullet$ \textbf{Question-aware Graph Transformer} encodes each type of MIS node and does reasoning among these nodes according to the question.

\noindent $\bullet$ \textbf{Answer Generator} generates the answer by incorporating question and updated graph nodes.

\subsection{MIS Graph Construction}

We first use the information retrieval system to retrieve related MIS as the knowledge source and construct a heterogeneous graph using these different types of MIS.
We take each type of MIS as a different type of node, and the question as another type of node.
Thus, we have five different types of nodes: (1) question; (2) user articles; (3) article comments; and (4) related questions and (5) its answers.
Intuitively, we connect all types of MIS documents with four basic edges: (1) user article to question; (2) related question to question; (3) related answer to related question; (4) comment to user article.

\subsection{Node Encoder}\label{sec:node-init}

Next, we employ the pretrained BART model to encode each MIS document independently:
\begin{align}
   \{{\bf h}^i_{j,1}, \cdots, {\bf h}^i_{j,L_d}\} &= \text{Enc}({d^i_{j,1}, \cdots, d^i_{j,L_d}}), \\
   \{{\bf h}^q_{1}, \cdots, {\bf h}^q_{L_q}\} &= \text{Enc}(w^q_1, \cdots, w^q_{L_q}),
\end{align}
where $\text{Enc}$ is the encoder module in BART which outputs the vector representation ${\bf h}^i_{j,k}$ of each input token $d^i_{j,k}$ in $j$-th document of MIS type $i$.
To obtain a vector representation of each MIS document, we apply the mean-pooling operation to the hidden states of tokens in a document:
\begin{equation}
    {\bf h}^i_j = \text{Mean-pooling} \left( \{{{\bf h}^i_{j,1}, \cdots, {\bf h}^i_{j,L_d}}\} \right). 
\end{equation}
We use the document representation ${\bf h}^i_j$ as the initial node representation of the heterogeneous graph.

\subsection{Question-aware Graph Transformer}

The GNN (Graph Neural Network) has been proven to offer better performance for the reasoning task in QA~\cite{Fang2020Hierarchical,Jiang2020Reasoning,Yasunaga2021QAGNN}.
The graph structure allows messages to pass over the nodes with local contextual information, which can comprehensively consider clue words in a different document to achieve the purpose of reasoning.
Previous works in QA show that GNNs have achieved promising results in reasoning tasks~\cite{Lan2020Query,Lv2020GraphBased}. 
Inspired by the heterogeneous graph transformer~\cite{Hu2020Heterogeneous}, in this paper, we propose a Question-aware Graph Transformer (QGT) which can aggregate question-related information from different type of MIS node by passing information among heterogeneous nodes.
An overview of the question-aware graph transformer is shown in the left part of Figure~\ref{fig:model}.

\subsubsection{Heterogeneous Question-aware Attention}

Since the structure information stored in the MIS graph is important for capturing the relationship between different types of MIS, we should fuse the information in neighbor nodes into the target node representation.
However, the salience of each node is not the same, and it is unreasonable to simply add all the information of all neighbor nodes into the target node.
Therefore, we conduct the multi-head mechanism~\cite{Vaswani2017Attention} to calculate the relevance score between the target node and the neighbor nodes.
We first project the input node representation into two spaces using different parameters for different type of MIS node:
\begin{align}
    V(s) = \text{MLP}_{\gamma(s)} \left( D[s] \right), \\
    K(t) = \text{MLP}_{\gamma(t)} \left( D[t] \right),
\end{align}
where $\gamma(\cdot)$ denotes the type of node, $D[\cdot]$ denotes the original node representation, $s$ denotes the index of one source node and $t$ denotes the index of the target node.
Next, we employ a bi-linear function to calculate the relevance between these nodes as the attention scores, and we also incorporate the edge type into the attention score calculation:
\begin{equation}
    \alpha(s, e, t) = \underset{\forall s \in N(t)}{\operatorname{Softmax}} \left( K(t) W_{\delta(e)} V(s) \right), \label{equ:attention-score}
\end{equation}
where $e$ is the edge between the node $s$ and node $t$, $N(t)$ is the neighbour nodes of node $t$, and the $W_{\delta(e)}$ is the edge type specific parameter.

\subsubsection{Heterogeneous Message Passing}

Intuitively, the message passing behavior on different edge types should not be the same.
For example, when passing a message between two user articles connected by an edge with the same entity words, we should pass the facts about the entity.
And the message between the related question and its answer should be the extracted facts from the answer according to the related answer.
Thus, the passed message from one node to other target nodes should not be the same when passing through a different type of edge.
We need to extract the useful message from the source node by considering the edge type:
\begin{equation}
    M(s, e, t) = \text{MLP}^{\text{msg}}_{\gamma(s)} \left( D[s] \right) W^{\text{msg}}_{\delta(e)}, \label{equ:message-pass}
\end{equation}
where $W^{\text{msg}}_{\delta(e)}$ is the edge type specific parameter matrix, and $M(s, e, t)$ is the message of source node $s$ that will be passed through edge $e$ to the target node $t$.

\subsubsection{Question-aware Aggregation}

Since the aim of conducting the message passing between nodes is to extract useful knowledge from heterogeneous MIS to answer the question, only the question-related knowledge should be reserved when passing the message.
Thus, first use the question relation score to re-scale the attention scores (calculated in Equation~\ref{equ:attention-score}):
\begin{equation}
    \hat{\alpha}(s, e, t) = \alpha(s, e, t) \times \beta(s),
\end{equation}
where $\beta(s) \in \mathbb{R}$ is the score for node $s$ which represents the semantic relevance between node $s$ and question $q$ calculated by a bi-linear layer:
\begin{equation}
    \beta(s) = D[q] W^r D[s],
\end{equation}
where $W^r$ is a trainable parameter, and the $D[q]$ and $D[s]$ are node representations for question $q$ and source node $s$ respectively.
Then, we apply a weighted sum over all the messages passed from all the source nodes:
\begin{equation}
    \tilde{D}[t] = \sum_{\forall s \in N(t)} \left( \hat{\alpha}(s, e, t) \cdot M(s, e, t) \right).
\end{equation}
Finally, we apply a linear projection with non-linear activation and a residual connect on the source nodes messages:
\begin{equation}
    \hat{D}[t] = \text{gelu} \left(\text{MLP}_{\gamma(t)} \tilde{D}[t] \right) + D[t] ,
\end{equation}
where $\hat{D}[t]$ is the updated node representation for target node $t$.
In this way, we obtain the contextualized vector representation of the target node which fuses the question-related information from neighborhood MIS nodes.

Inspired by the multi-layer mechanism of Transformer~\cite{Vaswani2017Attention}, we also conduct this question-aware graph transformer several times.
In each layer, the input is the output of the previous layer and the output will be used in the next layer as input.
The first layer uses the initial node representation as the input which is illustrated in \S~\ref{sec:node-init}.
After using the multi-layer heterogeneous graph transformer to encode the target node in the graph, we obtain the final vector representation of the target node.
For brevity, we omit the layer notion in our notion and equations, and use ${\bf g}^i_j = \hat{D}[t]$ as the notion of final node representation for the node $t$ which is the $j$-th node in $i$-th MIS type.

\subsection{Answer Generator}

HeteroQA generates the answer based on the knowledge learned from the MIS.
We employ the pretrained BART decoder as our base framework.
However, generating a good answer should incorporate the input question and the knowledge from MIS simultaneously, and the original BART decoder can only attend to the hidden states produced by the BART encoder.
We introduce a new graph attention layer and insert this layer into the original BART decoder, and preserve other layers in the BART.
We first apply the self-attention on the masked output embeddings of previous decoding steps which is same as the Transformer~\cite{Vaswani2017Attention}, and we obtain the output ${\bf p}^s$ for each decoding step.
Then we use the output ${\bf p}^s$ to cross-attend to the question hidden states:
\begin{align}
    {\bf p}^q = \text{MHAtt}({\bf p}^s, \{{\bf h}^q_{1}, \cdots, {\bf h}^q_{L_q}\}), \label{equ:cross-attn}
\end{align}
where $\text{MHAtt}$ is the standard multi-head attention layer~\cite{Vaswani2017Attention} and this procedure is the same as the original BART decoder.
After the self-attention and cross-attention layer, we apply a multi-head attention layer which aggregate useful knowledge from the updated graph nodes according to the state of each decoding step:
\begin{equation}
    {\bf p}^g = \text{MHAtt}({\bf p}^s, \{{\bf g}^1_{1}, \cdots, {\bf g}^T_{L^T_u}\}). \label{equ:graph-attn}
\end{equation}
Finally, we combine the question and knowledge from MIS:
\begin{equation}
    {\bf p}^f = {\bf p}^q + {\bf p}^g .
\end{equation}
We use the fused representation ${\bf p}^f$ to predict the distribution over the vocabulary of the generated answer, and use the cross-entropy loss $\mathcal{L}_e$ as the training objective to optimize the parameters.
 
\subsubsection{Loss Function}
 
To facilitate the question-aware graph transformer to capture the semantic relationship between each retrieved MIS and question, we propose an auxiliary question-aware loss function.
In detail, we employ a linear layer that uses the updated node representation as input to predict the BM25 score of each MIS node, and we calculate the mean squared error between the predicted score and the BM25 score produced by the information retrieval system as an auxiliary loss $\mathcal{L}_q$.
Finally, we combine text generation cross-entropy loss $\mathcal{L}_e$ and question-aware loss $\mathcal{L}_q$ as the final loss function $\mathcal{L} = \mathcal{L}_e + \psi \mathcal{L}_q$ where $\psi$ is a hyper-parameter.

%% file: exp.tex
\section{Experimental Setup}\label{sec:exp-setup}

\subsection{Research questions}

We list four research questions that guide the remainder of the paper:
\begin{enumerate}
  \item \textbf{RQ1} (See \S\ref{sec:overall-exp}): What is the overall performance of HeteroQA?
  \item \textbf{RQ2} (See \S\ref{sec:ablation-exp}): What is the effect of each module in HeteroQA?
  \item \textbf{RQ3} (See \S\ref{sec:doc-type-exp}): 
  Does every type of MIS help the CQA model to generate better answer?
  \item \textbf{RQ4} (See \S\ref{sec:doc-num-exp}): What is the effect of using more or less MIS in each type?
\end{enumerate}

\begin{table*}[t]
\begin{center}
\caption{Automatic metrics comparison between baselines.}
\label{tab:comp_bleu_baselines}
\resizebox{2.1\columnwidth}{!}{
\LARGE
\begin{tabular}{c|ccccccc|ccccccc}
\toprule
 & \multicolumn{7}{c}{AntQA} & \multicolumn{7}{|c}{$\text{MSM}^{\text{plus}}$} \\
\midrule
Method & BLEU & BLEU1 & BLEU2 & BLEU3 & BLEU4 & ROUGE-L & METEOR & BLEU & BLEU1 & BLEU2 & BLEU3 & BLEU4 & ROUGE-L & METEOR \\
\midrule
Retrieved1 & 1.11  & 8.35  & 1.95  & 0.52  & 0.18  & 7.32  & 0.15 & 0.21 & 4.82  & 0.25 & 0.04 & 0.03 & 3.78  & 0.07 \\
S2SA       & 13.40 & 24.54 & 14.01 & 10.57 & 8.88  & 19.42 & 0.18 & 0.39 & 7.36  & 0.82 & 0.13 & 0.03 & 10.08 & 0.06 \\
SASAU      & 14.80 & 25.14 & 15.14 & 11.98 & 10.53 & 19.73 & 0.18 & 0.57 & 8.60  & 1.09 & 0.20 & 0.05 & 11.23 & 0.07 \\
CVAE       & 13.15 & 23.10 & 13.54 & 10.55 & 9.07  & 18.44 & 0.17 & 0.38 & 7.45  & 0.80 & 0.13 & 0.03 & 10.37 & 0.05 \\ 
OAAG       & 2.19  & 3.77  & 2.32  & 1.77  & 1.49  & 17.07 & 0.13 & 1.56 & 8.61  & 1.25 & 0.78 & 0.70 & 8.49  & 0.06 \\
BART       & 16.47 & 27.93 & 16.99 & 13.37 & 11.61 & 22.06 & 0.21 & 6.66 & 21.06 & 8.28 & 4.34 & 2.60 & 28.82 & 0.19 \\ 
RAG        & 17.33 & 29.04 & 17.93 & 14.11 & 12.27 & 22.51 & 0.21 & 6.95 & 21.58 & 8.76 & 4.60 & 2.68 & 29.47 & 0.23 \\ \midrule 
HeteroQA        & \bf 19.71 & \bf 31.85 & \bf 20.30 & \bf 16.28 & \bf 14.35 & \bf 22.78 & \bf 0.22 & \bf 9.25 & \bf 24.17 & \bf 11.18 & \bf 6.50 & \bf 4.17 & \bf 34.63 & \bf 0.28 \\ 
HeteroQA-QGT    & 18.31 & 29.97 & 18.83 & 15.04 & 13.25 & 21.84 & 0.20 & 7.18 & 21.10 & 8.85 & 4.83 & 2.95 & 32.90 & 0.26 \\ 
HeteroQA-GL     & 18.90 & 30.80 & 19.46 & 15.53 & 13.71 & 22.26 & 0.20 & 7.20 & 21.06 & 8.85 & 4.85 & 2.97 & 32.96 & 0.26 \\ 
HeteroQA-HG     & 18.20 & 30.29 & 18.79 & 14.87 & 12.98 & 21.78 & 0.20 & 5.36 & 17.93 & 6.69 & 3.42 & 2.01 & 28.40 & 0.22 \\ 
\bottomrule
\end{tabular}
}
\end{center}
\end{table*}

\subsection{Evaluation Metrics}

To evaluate the methods, following existing text generation methods~\cite{Gao2019Abstractive,Gao2019How,Gao2020From}, we employ the automatic word overlap-based BLEU~\cite{Papineni2002BleuAM}, METEOR~\cite{banarjee2005} and ROUGE-L~\cite{lin2004rouge} to measure the lexical unit overlapping (\eg unigram, bigram) between the generated answer and ground truth.
Since only using automatic evaluation metrics can be misleading~\cite{Stent2005EvaluatingEM}, we also conduct the human evaluation. 
Three well-educated and full-time professional annotators are invited to judge 200 randomly sampled answers. 
The statistical significance of two runs is tested using a two-tailed paired t-test and is denoted using \dubbelop (or \dubbelneer) for strong significance for $\alpha = 0.01$.

\subsection{Comparisons}

To prove the effectiveness of each module, we conduct ablation studies which remove each key module in HeteroQA, and then form three baseline methods (1) \texttt{HeteroQA-QGT} uses the original HGT~\cite{Hu2020Heterogeneous} instead of our proposed question-graph transformer; (2) \texttt{HeteroQA-GL} removes the graph auxiliary loss $\mathcal{L}_q$; (3) \texttt{HeteroQA-HG} uses the homogeneous graph model GAT~\cite{Velickovic2018Graph} as the graph encoder.
Apart from the ablation study, we also compare with the following baselines: 

\noindent $\bullet$ \texttt{Retrieved1} uses the top-1 retrieved document as the answer for the user question.

\noindent $\bullet$ \texttt{S2SA} is the Sequence-to-Sequence framework~\cite{Sutskever2014SequenceTS} which is equipped with the attention mechanism~\cite{Bahdanau2015NeuralMT} and copy mechanism~\cite{Gu2016IncorporatingCM,See2017Get} as a baseline method. 
The input sequence is a question and the ground truth output sequence is the answer.
  
\noindent $\bullet$ \texttt{S2SAU} is a simple method that can incorporate the MIS information when generating the answer, we concatenate all types of MIS after the question as the input of \texttt{S2SA}.

\noindent $\bullet$ \texttt{CVAE}~\cite{Zhao2017Learning} is a text generation framework based on conditional variational autoencoder that uses latent variables to generate diverse sentences, and this framework has been applied in many knowledge-grounded text generation tasks~\cite{Kim2020Sequential,Lian2019Learning}.

\noindent $\bullet$ \texttt{BART}~\cite{Lewis2020BART} is a denoising autoencoder for pretraining sequence-to-sequence models, and it is particularly effective when fine-tuned for text generation but also works well for comprehension tasks.

\noindent $\bullet$ \texttt{RAG}~\cite{Lewis2020RetrievalAugmented} is general-purpose fine-tuning recipe for retrieval-augmented generation models which combine pretrained non-parametric memory for knowledge-intensive NLP tasks. 
 \texttt{RAG} outperforms many strong Open-Domain QA and CQA methods. 

\noindent $\bullet$ \texttt{OAAG}~\cite{Deng2020Opinion} is a generative community question answering model which jointly models opinionated and interrelated information between the question and contextual document.

\subsection{Implementation Details}

We implement our experiments using PyTorch~\cite{Paszke2019PyTorchAI} based on the Transformers~\cite{wolf-etal-2020-transformers}.
We train our model on two NVIDIA V100 GPU for four days. 
We employ the pretrained BART-base model (with 6 layers for encoder and decoder, the number of attention head is 12, and the hidden size is 768) to initialize part of the parameters.
The hyper-parameter $\psi$ is set to 0.01.
Since there is not a public pretrained Chinese BART, we pretrain the BART-base from scratch using the text infilling objective on 7GB Chinese text data, and train the model for two weeks using 8 NVIDIA V100 GPU.

\newcommand{\cbkgrnd}{\cellcolor{blue!15}}
\newcommand{\sbbkgrnd}{\cellcolor{gray!65}}
\newcommand{\phantomtriangle}{\phantom{\dubbelop}}

\section{Experimental Results} \label{sec:exp-result}

\begin{table}[t]
\begin{center}
\caption{Human Evaluation Results.}
\label{tab:comp_human}
\resizebox{0.7\columnwidth}{!}{
\begin{tabular}{c|c|cc}
\toprule
Dataset & Method & Fluency & Correctness \\
\midrule

\multirow{4}*{\rotatebox{90}{AntQA}}
& OAAG & 2.49 & 1.84 \\ 
& BART & 2.58 & 1.96 \\ 
& \cbkgrnd RAG & \cbkgrnd 2.44 & \cbkgrnd 2.11 \\ \cline{2-4}
& HeteroQA  & \bf 2.61\dubbelop & \bf 2.25\dubbelop \\ 
\midrule

\multirow{4}*{\rotatebox{90}{$\text{MSM}^{\text{plus}}$}}
& OAAG & 2.68 & 1.56 \\ 
& BART & 2.89 & 1.65 \\ 
& \cbkgrnd RAG  & \cbkgrnd 2.88 & \cbkgrnd 1.55 \\ \cline{2-4}
& HeteroQA  & \bf 2.90\dubbelop & \bf 1.83\dubbelop \\ 
\bottomrule
\end{tabular}
}
\end{center}
\end{table}

\subsection{Overall Performance}\label{sec:overall-exp}

We compare our model with the baselines listed in Table~\ref{tab:comp_bleu_baselines}.
Our model performs consistently better on two datasets than other CQA and conditional text generation models with improvements of 13.73\%, 9.68\%, and 13.22\% on the AntQA dataset, and achieves 33.09\%, 12.00\% and 27.63\% improvements on the $\text{MSM}^{\text{plus}}$ dataset compared with RAG in terms of BLEU, BLEU1, and BLEU2 respectively.
This demonstrates that MIS in the user community provides useful knowledge for answer generation that cannot be replaced by other complicated baselines without this information.
And our HeteroQA method has a greater performance improvement compared to the methods that do not use the pretrained language model.
\citet{Sun2020EndCold} also propose a retrieval-based CQA method, which leverages heterogeneous graph to route the cold question to the exiting QA nodes in the CQA graph and they did not incorporate the MIS in the user community.
Since they did not release their source code, we use a simple and intuitive method \texttt{Retrieved1} to verify the basic performance of these retrieval-based methods.
From Table~\ref{tab:comp_bleu_baselines} we can find that the retrieval-based method achieves the worse performance among all the baselines, which demonstrates the retrieval-based method is not suitable for the generative CQA.

For the human evaluation, we asked annotators to rate each summary according to its factual correctness and language fluency.
The rating score ranges from 1 to 3, with 3 being the best.
Table~\ref{tab:comp_human} lists the average scores of each model, showing that HeteroQA outperforms the other baseline models in both fluency and correctness.
To prove the significance of these results, we also conduct the paired student t-test between our model and \texttt{RAG} and obtain $p < 0.05$ for fluency and consistency, respectively.
From this experiment, we find that the \texttt{BART} achieves a comparable fluency score as HeteroQA, the correctness score is significantly lower than HeteroQA.
We also find that there are 22\% samples generated by our model obtain a low score (1) for correctness.
To verify the reason for generating these bad cases, we find that in the input MIS of these bad cases, there are 86.36\% samples that obtain a score 1 for the quality of MIS.
This phenomenon demonstrates that the misleading MIS input leads our HeteroQA model to generate an answer with wrong facts.

Although our HeteroQA model achieves 13.73\% improvement in terms of BLEU over \texttt{RAG} model on AntQA, the increment is smaller than the performance on the $\text{MSM}^{\text{plus}}$ dataset which achieves 33.09\% improvement in terms of BLEU.
The reason is that AntQA is more similar to the real-world scenario where the supported MIS are all retrieved by an information retrieval system using the BM25 algorithm and we did not employ any labor-intensive human annotation to filter the data which contains some noisy samples.
Conversely, the supporting document in $\text{MSM}^{\text{plus}}$ is filtered by a human annotator, which has less noisy data in their dataset.

\subsection{Ablation Studies}\label{sec:ablation-exp}

The performance of different ablation models are shown in Table~\ref{tab:comp_bleu_baselines}.
All ablation models perform worse than HeteroQA in terms of all metrics, which demonstrates the preeminence of HeteroQA. 
The experiment shows that our HeteroQA achieves 7.65\% and 28.83\% increments compared with \texttt{HeteroQA-QGT} in terms of BLEU score on AntQA and $\text{MSM}^{\text{plus}}$ respectively, which verifies the effectiveness of our proposed QGT.
In the method \texttt{HeteroQA-HG}, we employ the GAT~\cite{Velickovic2018Graph} as the graph encoder and there is a slight decrement from \texttt{HeteroQA-HG} to \texttt{HeteroQA-QGT} of 0.28\% and 15.85\% in terms of ROUGE-L on AntQA and $\text{MSM}^{\text{plus}}$, which demonstrates the effectiveness of treating each type of MIS as different nodes and using the heterogeneous graph to model these MIS.
Thus, we can verify the contributions of each module in HeteroQA.

\subsection{Effect of Different Type of MIS}\label{sec:doc-type-exp}

\begin{table}[t]
\begin{center}
\caption{Effective of using different type of MIS.}
\label{tab:comp_bleu_ugc_ablations}
\resizebox{0.8\columnwidth}{!}{
\begin{tabular}{c|c|ccc}
\toprule
Dataset & Method & BLEU & ROUGE-L & METEOR \\
\midrule

\multirow{5}*{\rotatebox{90}{AntQA}}
& HeteroQA     & 19.71 & 22.78 & 0.22 \\ \cline{2-5}
& HeteroQA-QA  & 17.76 & 20.95 & 0.20 \\ 
& HeteroQA-COM & 18.56 & 21.81 & 0.21 \\
& HeteroQA-ART & 17.91 & 21.55 & 0.19 \\
\midrule

\multirow{5}*{\rotatebox{90}{$\text{MSM}^{\text{plus}}$}}
& HeteroQA     & 9.25 & 34.63 & 0.28 \\ \cline{2-5} 
& HeteroQA-QA  & 7.08 & 32.73 & 0.25 \\ 
& HeteroQA-COM &   -  &   -   & -    \\
& HeteroQA-ART & 5.31 & 28.47 & 0.22 \\
\bottomrule
\end{tabular}
}
\end{center}
\end{table}

To verify the effectiveness of each MIS type, we remove each type of MIS from our heterogeneous graph. 
We employ three experiments that remove the MIS type one-by-one: \texttt{HeteroQA-QA} (w/ comment, article), \texttt{HeteroQA-COM} (w/ article, related question and answer) and \texttt{HeteroQA-ART} (w/ related question and answer).
Since the comment of the documents in $\text{MSM}^{\text{plus}}$ can not be obtained, we omit this experiment.
Table~\ref{tab:comp_bleu_ugc_ablations} shows the performance of these ablation models, and we only show three metrics of each model due to the limited space.
From this table, we find that the related QA produces the greatest contribution (10.98\% decline in terms of the BLEU) on AntQA, and the related articles produce the greatest contribution on $\text{MSM}^{\text{plus}}$ with a 74.20\% decline in terms of the BLEU.
Although the contribution of using comment of articles is minimum compared to other MIS types, it also increases the BLEU score for 6.20\%.

\subsection{Effect of Using Different Number of MIS}\label{sec:doc-num-exp}

\begin{figure} 
    \centering
    \subfigure{ 
      \includegraphics[width=0.47\columnwidth]{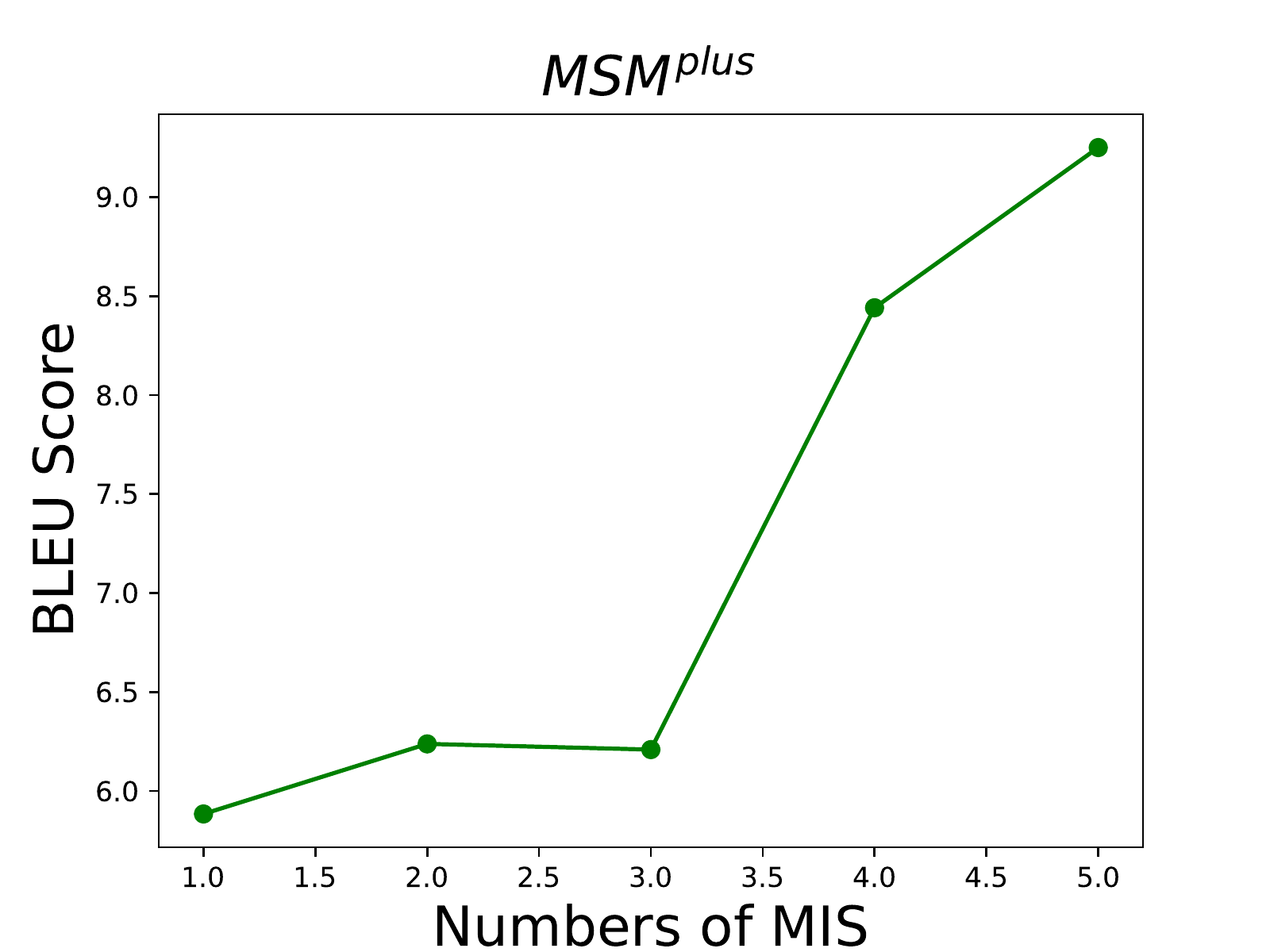}
    }
    \subfigure{ 
      \includegraphics[width=0.47\columnwidth]{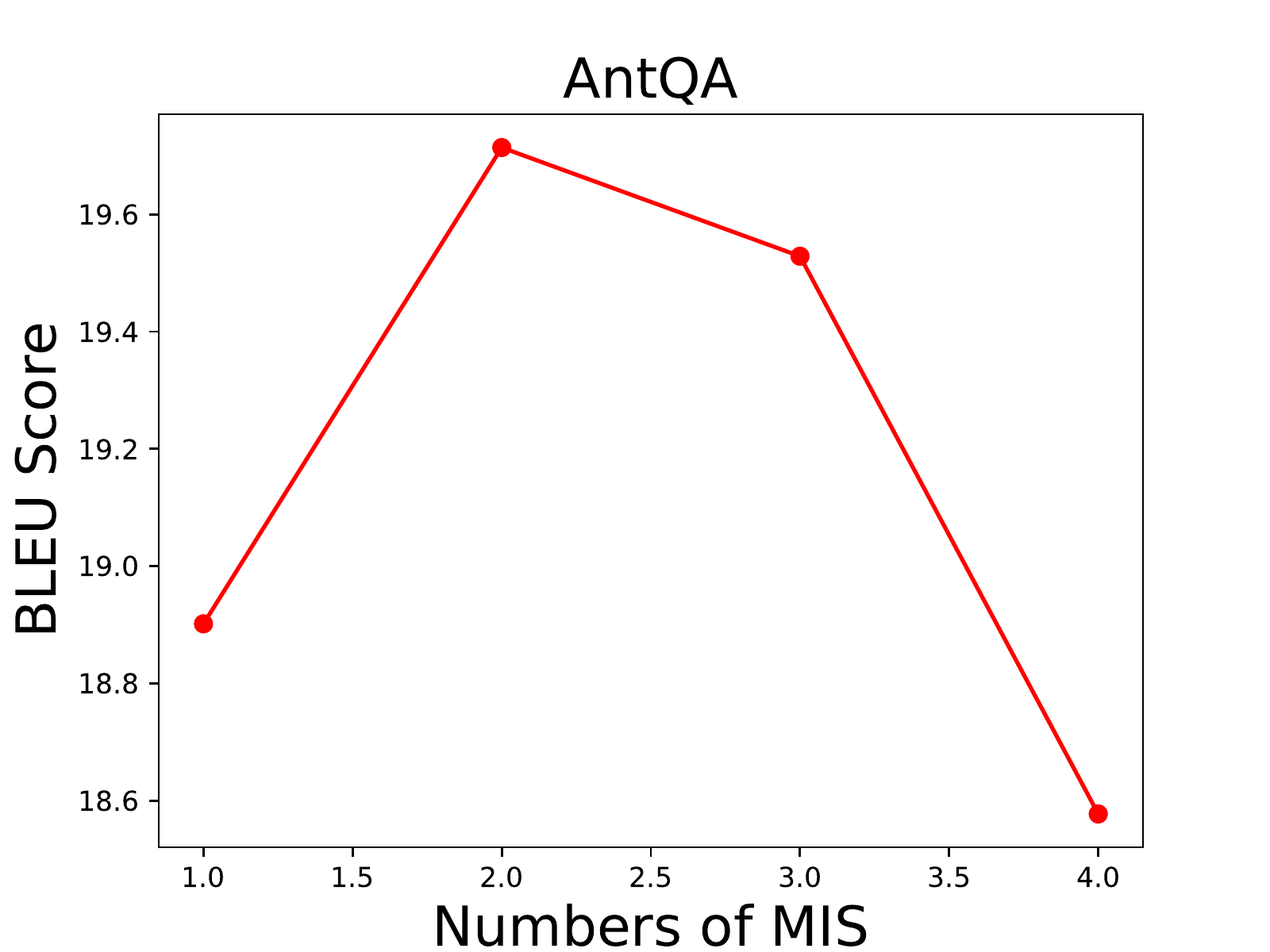}
    }
    \caption{Visualization of BLEU score with different MIS document number.}
    \label{fig:ugc-num}
\end{figure}

We investigate the influence of the document number for each MIS type incorporated by the HeteroQA model.
Figure~\ref{fig:ugc-num} illustrates the relationship between the document number in each type of MIS and the BLEU score.
On the $\text{MSM}^{\text{plus}}$ dataset, the performance keeps growing when the number of document reaches 5, and due to the hardware limitation, we did not conduct experiments for more MIS.
Since the document in $\text{MSM}^{\text{plus}}$ is filtered by humans and all the related articles can support answering the question, the performance of the model can keep increasing when the model reads more MIS.
On the AntQA dataset, the BLEU score first increases with the number of MIS, and after reaching an upper limit it then begins to drop.
This phenomenon demonstrates that incorporating several relevant MIS can help the model to achieve better performance.
In the real-world application, we cannot filter the contextual information case-by-case.
And excessive MIS which contains not relevant information will confuse the model to extract useful knowledge to generate correct answers.

\subsection{Case Study}

\input{case}

Table~\ref{tab:case} shows an example of the generated answer and MIS context with useful knowledge to answer the user question.
Although all the methods generate fluent answers, the facts described in \texttt{BART} and \texttt{RAG} are wrong (the wrong facts are shown in \textcolor{red}{red text} in Table~\ref{tab:case}).
And HeteroQA generates refined and correct repayment steps for the user question according to the related QA and user article.

%% file: case.tex
\begin{CJK*}{UTF8}{gbsn}
    \begin{table}[t]
    \centering
    \caption{Examples of the context and answers. We only show two MIS due to the limited space. We use \textcolor{red}{red text} to indicate wrong facts in generated answers.}
    \label{tab:case}
    \resizebox{1\columnwidth}{!}{
    \begin{tabular}{l}
    \toprule
    \multicolumn{1}{p{1\columnwidth}}{\textbf{Question}: 在支付宝上借款怎么还款 (How to conduct repayment on Alipay?)} \\
    \multicolumn{1}{p{1\columnwidth}}{\textbf{MIS (Related QA)}: \textbf{Q}: 支付宝借呗怎么还？(How to repay for ``Jiebei'' (Ant Microloan)?) \textbf{A}: 进入支付宝借呗页面，就可以看到去借钱和去还钱，点击还款就可以看到还款数目，将足够的资金提前存入支付宝的余额 (Login the ``Jiebei'' page, and you can see ``Borrow money'' and ``Repayment''. Click on the ``Repayment'' button, and then you can see the amount you should pay back. Deposit enough money in your Alipay balance account.)} \\
    \multicolumn{1}{p{1\columnwidth}}{\textbf{MIS (Related Article)}: 借呗提前还款要收手续费吗？不需要，借呗提前还款暂时没有手续费，大家直接在借呗页面操作即可。登录手机支付宝，点击右下角，在借呗首页选择，按页面提示操作。(Do I have to charge a handling fee for early repayment? No, there is no handling fee for repayment in ``Jiebei''. So you can directly conduct repayment on the ``Jiebei''. First, log in to Alipay mobile version, click on the lower right corner, select on the home page of ``Jiebei'', and follow the instructions on the page.)} \\
    \multicolumn{1}{p{1\columnwidth}}{\textbf{BART}: 登录支付宝，点击我的，\textcolor{red}{点击借款}，点击还款就可以了 (Login Alipay, then click on ``My Apps'', \textcolor{red}{then click on ``Borrow some money''}, then click on ``Repayment'' and that's it.)} \\
    \multicolumn{1}{p{1\columnwidth}}{\textbf{RAG}: 支付宝借呗\textcolor{red}{借款的步骤}，打开支付宝钱包，点击借呗，然后点击还款，\textcolor{red}{选择借款方式}，就可以还款了 (The \textcolor{red}{steps of borrowing money} from ``Jiebei'' (Ant Microloan). Login ``Pocket'' and click on ``Jiebei''. Then click on ``Repayment'' and \textcolor{red}{select the way of making a loan}.)} \\
    \multicolumn{1}{p{1\columnwidth}}{\textbf{HAG}: 进入支付宝，点击借呗，然后点击还款就可以了 (Login Alipay, click on ``Jiebei'' (Ant Microloan), and then click on ``Repayment''.)} \\
    \bottomrule
    \end{tabular}
    }
    \end{table}
\end{CJK*}

%% file: conclusion.tex
\section{Conclusion} \label{sec:conclusion}

In this paper, we propose the Heterogeneous graph-based Question Answering model (HeteroQA) to tackle the community question answering task, which extracts useful knowledge from Multiple Information Sources (MIS) which contains four types documents in the user community.
Since MIS have different contexts and structures, we treat MIS as heterogeneous information and propose a question-aware heterogeneous graph transformer to model the relationship between these nodes.
To generate correct and fluent answer text, we leverage the pretrain language model as an answer generator and introduce a graph attention layer to fuse the knowledge into the answer generation.
Experiments conducted on two datasets verify the effectiveness of our HeteroQA model.